\documentclass[conference]{IEEEtran}
\IEEEoverridecommandlockouts

\usepackage{cite}
\usepackage{amsmath,amssymb,amsfonts}
\usepackage{algorithmic}
\usepackage{graphicx}
\usepackage{textcomp}
\usepackage{xcolor}

\usepackage{tikz}
\usepackage{caption}
\usepackage{url}
\usepackage{array}
\usepackage{tabularx}
\usepackage{microtype}
\usepackage{hyperref}

\usepackage[ruled,vlined,linesnumbered]{algorithm2e}

\usepackage{multirow}
\usepackage{makecell}
\usepackage{booktabs}
\usepackage{etoolbox}

\def\BibTeX{{\rm B\kern-.05em{\sc i\kern-.025em b}\kern-.08em
    T\kern-.1667em\lower.7ex\hbox{E}\kern-.125emX}}

\usepackage[small,compact]{titlesec} 
\titlespacing{\section}{2.5pt}{2.5pt}{2.5pt}
\titlespacing{\subsection}{2pt}{2pt}{2pt}
\titlespacing{\subsubsection}{2pt}{2pt}{2pt}

\makeatletter
\newcommand{\linebreakand}{%
  \end{@IEEEauthorhalign}
  \hfill\mbox{}\par
  \mbox{}\hfill\begin{@IEEEauthorhalign}
}
\makeatother

\begin{document}

\title{OptINC: Optical In-Network-Computing for Scalable Distributed Learning}

\author{
\IEEEauthorblockN{1\textsuperscript{st} Sijie Fei}
\IEEEauthorblockA{\textit{Technical University of Munich} \\
Munich, Germany \\
sijie.fei@tum.de}
\and
\IEEEauthorblockN{2\textsuperscript{nd} Grace Li Zhang}
\IEEEauthorblockA{\textit{Technical University of Darmstadt} \\
Darmstadt, Germany \\
grace.zhang@tu-darmstadt.de}
\and
\IEEEauthorblockN{3\textsuperscript{rd} Bing Li}
\IEEEauthorblockA{\textit{Technical University of Ilmenau} \\
Ilmenau, Germany \\
bing.li@tu-ilmenau.de}
\linebreakand
\IEEEauthorblockN{4\textsuperscript{th} Ulf Schlichtmann}
\IEEEauthorblockA{\textit{Technical University of Munich} \\
Munich, Germany \\
ulf.schlichtmann@tum.de}
}

\maketitle
\begin{abstract}
Distributed learning is widely used for training large models on large datasets by distributing parts of the model or dataset across multiple devices and aggregating the computed results for subsequent computations or parameter updates. Existing communication algorithms for distributed learning such as ring all-reduce result in heavy communication overhead between servers. Since communication in large-scale systems uses optical fibers, we propose an Optical In-Network-Computing (OptINC) architecture to offload the computation in servers onto the optical interconnects. To execute gradient averaging and quantization in the optical domain, we incorporate optical devices such as Mach-Zehnder-Interferometers (MZIs) into the interconnects. Such a de facto optical neural network (ONN) can effectively reduce the communication overhead in existing distributed training solutions. To reduce dataset complexity for training this neural network, a preprocessing algorithm implemented in the optical domain is also proposed. Hardware cost is lowered by approximating the weight matrices of the optical neural network with unitary and diagonal matrices, while the accuracy is maintained by a proposed hardware-aware training algorithm. The proposed solution was evaluated on real distributed learning tasks, including ResNet50 on CIFAR-100, and a LLaMA-based network on Wikipedia-1B. In both cases, the proposed framework can achieve comparable training accuracy to the ring all-reduce baseline, while eliminating communication overhead.
\end{abstract}

\section{Introduction}
\label{inrtoduction}
 In recent years, deep neural networks (DNNs) have achieved remarkable progress in various scenarios such as Computer Vision (CV) and Natural Language Processing (NLP). To address increasingly complex tasks, DNNs, especially Large Language Models (LLMs) in the NLP domain, are growing rapidly in both model sizes and training data volume. For example, state-of-the-art LLMs can contain hundreds of billions of parameters and are trained on trillions of tokens~\cite{llama2,gpt3,basissharing}. The increasing model parameters and training data are posing significant challenges for training on a single GPU due to memory and computational limitations.   


To overcome these limitations, distributed learning is applied to train a single model across multiple GPUs or servers. Generally, the model is trained via model parallelism~\cite{megatronlm,pipeline_parallelism}, where model parameters are split across devices, or data parallelism, where each device trains on a local data subset. For both strategies, the resulting partial outputs or gradients are aggregated and processed after every batch. These two strategies are orthogonal and can be combined~\cite{megatronlm,megatronlm2}.

\begin{figure}[!t]
\captionsetup{skip=8pt}
\centerline{\includegraphics[scale=0.65]{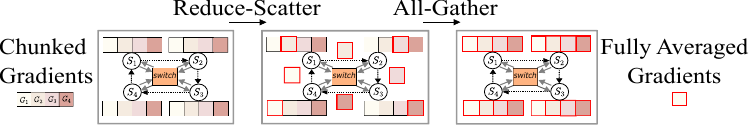}}
\small
\captionsetup{font=small}
\caption{\textbf{The ring all-reduce algorithm in distributed training with four servers connected to a switch, forming a logical ring topology. }}
\vspace{-20pt}
\label{ring}
\end{figure}

However, both strategies require frequent inter-device communication during the training. To reduce the communication, the ring all-reduce algorithm~\cite{ringallreduce} is widely used. Fig.~\ref{ring} shows an example of data parallelism with four servers forming a logical ring, where the gradients in the four servers should be averaged and synchronized. To balance the communication workload, the gradients in each server are partitioned into four chunks. The process consists of two stages. First, in the Reduce-Scatter stage, in every communication round, each server simultaneously sends one chunk to one neighbor and receives one chunk from another, where the averaging operation is performed in the servers. After three such rounds, each server holds one distinct chunk of the fully averaged gradients. Second, in the All-Gather stage, the servers redistribute the averaged chunks to others, requiring another three rounds.
In general, with $N$ servers, while transmitting all chunks theoretically requires only $N$ rounds, the ring all-reduce algorithm requires $2(N-1)$ rounds, resulting in a relative communication overhead of $\frac{N-2}{N}$, which is nearly 100\%. On modern GPUs, compute capabilities often outpace communication bandwidth~\cite{inc1}, making communication overhead a  bottleneck in large-scale distributed training.


To reduce this communication overhead, several strategies have been proposed. For instance, alternative logical topologies~\cite{treeallreduce,clos} are introduced, but they are often too complex to deploy at scale. Another direction is to quantize tensors and gradients to lower bit widths~\cite{deepcompression,qsgd,10137171}, which can reduce bandwidth but lead to model accuracy degradation. Despite these efforts, however, all these approaches still rely on multiple rounds of communication, which remain a key communication bottleneck.

To eliminate redundant communication rounds, In-Network Computing (INC) has been introduced, where the computations of tensors or gradients can be offloaded from each device to the network itself. Prior work ~\cite{inc1,inc2,inc3} has embedded computation and control units into electrical switches. When the data packets pass by the switches, they can be processed and aggregated directly in the switches. This can reduce latency significantly and accelerate training by 1.4x to 5.5x~\cite{inc1,inc2}. 

However, implementing INC on electrical switches introduces several drawbacks. First, performing INC on electrical switches can increase energy consumption due to optical-electrical-optical (O-E-O) conversions. Moreover, the partial results need to be buffered until computation in the electrical switches is complete, leading to packet evictions~\cite{inc1}. Alternatively, Optical Circuit Switch (OCS)~\cite{jupiter} can avoid the optical-electrical-optical conversions, but the potential computation ability in the OCS has not been explored.

In this paper, we propose a novel Optical INC (OptINC) architecture for data parallelism, which eliminates both communication overhead and optical-electrical-optical conversions. Our contributions are summarized as follows:

\begin{itemize}
\item 
An optical INC architecture with an Optical Neural Network (ONN) based on Mach-Zehnder-Interferometers (MZIs) is proposed to offload the computation in servers onto the optical interconnects, eliminating communication overheads. 

\item
To reduce the hardware cost, weight matrices in selected layers of the ONN are approximated by submatrices.
To maintain accuracy, a novel hardware-aware training scheme is adopted, recovering the ONN accuracy perfectly. 

\item 
The proposed OptINC architecture is shown to be scalable in a cascading manner, incurring small hardware overhead. 
Trained with a modified dataset, OptINC can efficiently eliminate quantized errors caused by multi-level quantization.
\end{itemize}

The rest of the paper is organized as follows: Section~\ref{preliminaries} provides background on OCS and MZIs. Section~\ref{proposed work} introduces the OptINC architecture and the hardware-aware training and adaptations made for scalable architectures. Experimental results are shown in Section~\ref{experimental results}. Section~\ref{conclusion} draws the conclusion.

\section{Preliminaries}
\label{preliminaries}
In this section, we first discuss why the OCS is suitable for HPC. Then we introduce how MZIs can implement the matrices in the OCS and ONNs. 

\subsection{OCS for HPC}
With the increasing demand for bandwidth, datacenters and HPC rely on optical fibers for communication. Since electrical switches require O-E-O conversions, which incur energy overhead and latency, optical switches provide more efficient communication by avoiding these conversions.

An OCS can be realized with Micro-Electro-Mechanical Systems (MEMS)~\cite{tpu,jupiter} with mechanically controlled mirrors or optical devices such as MZIs~\cite{mziocs}. 
In MZI-based OCS, reconfigurable connection matrices can be implemented to direct signals to output ports by tuning the phase shifters. 

Due to the slow tuning speed of mechanical mirrors and optical devices, a common challenge for the OCS is the reconfiguration latency, typically on the order of $\mu$s~\cite{psspeed} or even ms~\cite{mirrorspeed}. However, in distributed learning, the communication patterns are predetermined and require few changes during the training, making the reprogramming costs negligible. Therefore, with high bandwidth and no O-E-O conversions, the OCS is suitable for distributed learning tasks in HPC.

\subsection{Matrix-Vector Multiplications in OCS and ONNs}
\label{ONNs implemented with MZIs}


\begin{figure}[!t]
\centering
\captionsetup{skip=8pt}
\centerline{\includegraphics[scale=0.65]{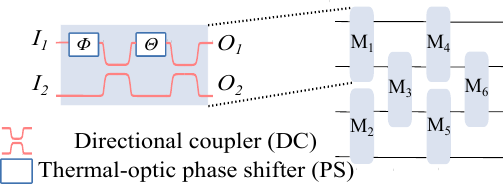}}
\small
\captionsetup{font=small}
\caption{\textbf{
The interleaving MZI array for a 4$\times$4 unitary matrix (adopted from~\cite{onn5}) where an MZI consists of two DCs and two PSs.
}}
\vspace{-20pt}
\label{mzi}
\end{figure}

To implement matrix-vector multiplication in neural networks, in-memory-computing with emerging devices such as RRAM~\cite{8714954, 9116244,9073995,10137106} and optical devices. 
Fig.~\ref{mzi} shows how MZIs can be used to build a connection matrix in the OCS. An MZI consists of two directional couplers and two thermal-optic phase shifters. The input information is encoded on the amplitudes of two optical signals, $I_1$ and $I_2$, which are input from the left ports of the MZI. Along the optical paths, the optical signals are transformed and output as $O_1$ and $O_2$ from the right ports. 
The transformation matrix is inherently unitary.




A unitary $M\times M$ matrix can be implemented by cascading $\frac{M\times (M-1)}{2}$ MZIs in an interleaving manner. For example, Fig.~\ref{mzi} shows how a 4$\times$4 matrix can be realized by six MZIs, $M_1$ to $M_6$. 
To implement an arbitrary $M\times N$ matrix $W$ with MZIs, $W$ is decomposed using Singular Value Decomposition (SVD) as follows:
\begin{equation}
    W \overset{\text{SVD}}{=} U\Sigma V^\top
\label{SVD}
\end{equation}
where $U$ and $V$ denote an $M\times M$ unitary matrix and an $N\times N$ unitary matrix, respectively, and can be implemented with cascading MZIs. $\Sigma$ is an $M\times N$ diagonal matrix and
can be implemented with a column of MZIs. In total, the implementation of $W$ requires $\frac{M(M+1)+N(N-1)}{2}$ MZIs. To realize a specific matrix, the PSs in the MZIs should be programmed to appropriate values by tuning the heaters~\cite{psspeed}. 

MZIs are widely used to implement linear weight matrices in ONNs~\cite{onn1,onn2,onn3,onn4,onn5}. The nonlinear activations can be realized either in digital circuits~\cite{onn1} with optical-electrical-optical conversions, or directly in the optical domain using electro-optic devices~\cite{nonlinear1} or nonlinear materials~\cite{nonlinear2}.

\section{Proposed work}
\label{proposed work}

To reduce the $\frac{N-2}{N}$ communication overhead in ring all-reduce for distributed learning, we propose an OptINC architecture for data parallelism that performs gradient averaging and quantization directly within the network, consisting of linear operations and nonlinear logic. Conventional optical logic gates usually require dedicated and specialized control mechanisms~\cite{optic_logic}, making them difficult to deploy for HPC. Instead, an ONN is employed to perform the computation. 


\subsection{OptINC Architecture}
\label{Optical INC Architecture}

\begin{figure}[!t]
\captionsetup{skip=5pt}
\centerline{\includegraphics[scale=0.9]{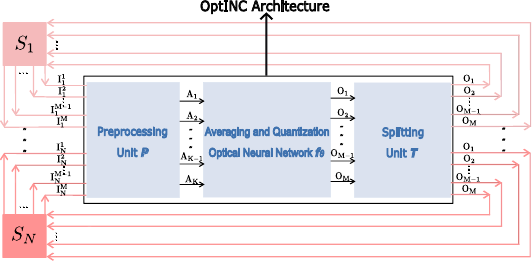}}
\small
\captionsetup{font=small}
\caption{\textbf{The proposed OptINC architecture connecting $N$ servers, $S_1$ to $S_N$, each with $M$ full-duplex optical transceivers. The system consists of three components: a preprocessing unit $\mathbf{P}$, an ONN $\boldsymbol{f}_\theta$, and a splitting unit $\mathbf{T}$.}}
\vspace{-20pt}
\label{archi}
\end{figure}

Fig.~\ref{archi} illustrates the proposed OptINC architecture supporting $N$ servers, $S_1$ to $S_N$. Unlike the logical ring topology in Fig.~\ref{ring}, all servers are just connected to the OptINC without forming additional logical topologies. Each server is equipped with $M$ full-duplex optical transceivers. As shown in Fig.~\ref{archi}, one server can send data $I_i$ to the network via the $i$-th transceiver, while receiving data $O_i$ from the architecture using the same transceiver simultaneously.
Before transmission, each server should encode its local gradient into optical signals. While gradients are typically stored in 16 or 32-bit formats~\cite{mixed}, due to transmission efficiency and reliability, HPC typically adopts 4-level Pulse Amplitude Modulation (PAM4)~\cite{pam4}. Therefore, a $B$-bit local gradient in the server $n$ , $G_n$, is encoded to $M=\lceil B/2 \rceil$ 2-bit segments, each of which is mapped to a PAM4 signal. The $i$-th PAM4 signal for server $n$, $I_n^{(i)}$, is extracted using the following operation:
\begin{equation}
I_n^{(i)} \;=\; \left\lfloor \frac{G_n}{2^{\,2(M-i)}} \right\rfloor \bmod 4,
\qquad i = 1, 2, \dots, M.
\label{encode}
\end{equation}

After transmission through the OptINC, each server decodes the optical signals and reconstructs the $B$-bit global quantized averaged gradient $\overline{G}$ using (\ref{encode}) inversely. Since the transceivers have limited resolution, the received optical signals are quantized back to the nearest PAM4 level by the transceivers.



The goal of the OptINC architecture is to execute gradient averaging and quantization directly within the network. By offloading the computation from the servers to the optical interconnects, communication overhead introduced by extra synchronization rounds as in the ring all-reduce algorithm can be eliminated. During the computation, although gradient averaging is a linear operation and can be performed by a linear matrix, quantization and the mapping of the averaged gradient to discrete PAM4 signals are nonlinear. 
Therefore, we employ an ONN with linear weight matrices implemented by the interleaving MZIs in Fig.~\ref{mzi} and nonlinear activation functions, denoted by $\boldsymbol{f}_\theta$ in Fig.~\ref{archi}.



The ONN $\boldsymbol{f}_\theta$ is used to approximate the quantized average of all local gradients. Ideally, the reconstructed gradient $\overline{G}$ from the received output signals
should match the expected quantized average gradient $\overline{G}^*$:
\begin{equation}
\overline{G} \stackrel{!}{=} \overline{G}^* = \mathbf{Q}( \frac{1}{N}\sum_{n=1}^{N} G_n )
\label{equation9}
\end{equation}
where $ \mathbf{Q}(\cdot)$ denotes the quantization and $G_n$ is defined in (\ref{encode}).


The gradient averaging, quantization, and the mapping function in the ONN
$\boldsymbol{f}_\theta$ need to process $M$ encoded PAM4 optical signals transmitted by each of the $N$ servers. 
Accordingly, the ONN needs to be trained to process $2^{MN}$ input combinations. As the number of servers $N$ grows, the dataset size for training the ONN grows exponentially. To lower the data complexity, a preprocessing unit is introduced before the ONN, as shown by $\mathbf{P}$ in Fig.~\ref{archi}, reducing the ONN input size to $K\leq M$, by 
%
%
averaging every $\lceil \frac{M}{K}\rceil$ signals from $N$ servers. 
The averaged input signal to the ONN is denoted as $A_k$ for $k=1,2\dots K$, as shown in Fig.~\ref{archi}. Since the sum of $\lceil \frac{M}{K}\rceil$ PAM4 signals ranges from 0 to $N(4^{\lceil \frac{M}{K}\rceil}-1)$, the averaged value across $N$ servers, namely $A_k$, ranges from 0 to $4^{\lceil \frac{M}{K}\rceil}-1$ with resolution $\frac{1}{N}$. With $K$ such inputs, the dataset size required for training is reduced to $(N(4^{\lceil \frac{M}{K}\rceil}-1)+1)^K$. As a result, the data complexity can be reduced from $O(2^{MN})$ to  $O(2^{K})$. $K$ is determined by balancing the data complexity and training efficiency.



Since all servers receive the same signals representing the averaged gradients, a simple signal splitting unit $\mathbf{T}$ is employed to broadcast the ONN outputs to each server, as shown in Fig.~\ref{archi}. This function can be implemented  using a simple MZI array.

\subsection{Hardware-Aware Design and Training of ONN \texorpdfstring{$\boldsymbol{f}_\theta$}{f(theta)}}
\label{training}

\begin{figure}[!t]
\captionsetup{skip=8pt}
\centerline{\includegraphics[scale=0.8]{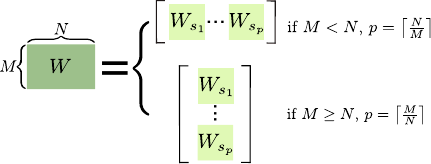}}
\small
\captionsetup{font=small}
\caption{\textbf{Weight matrix $W$ can be partitioned to square submatrices $W_s$ in two ways, horizontally or vertically. 
}}
\vspace{-20pt}
\label{matrix partition}
\end{figure}

As stated in Section~\ref{ONNs implemented with MZIs}, to implement an ONN, an $M\times N$ weight matrix $W$ requires $\frac{M(M+1)+N(N-1)}{2}$ MZIs. If one dimension is much larger than the other, the hardware area of the weight matrix will be dominated by the larger dimension. To reduce the area cost of a weight matrix, a matrix approximation approach similar to that in~\cite{onn4,onn3} is adopted. Specifically, $W$ is first partitioned into small square submatrices $W_s$ as shown in Fig.~\ref{matrix partition}, after which the partitioning can be performed either horizontally or vertically.

After partitioning, instead of using SVD in (\ref{SVD}), each square submatrix $W_s$ can be approximated with $W_a$, composed of only one diagonal matrix $\Sigma_a$ and one unitary matrix $U_a$: 
\begin{equation}
    W_s \approx W_a = \Sigma_a U_a
\label{approxi1}
\end{equation}
\begin{equation}
    U_a = U_sV_s^\top
\label{approxi2}
\end{equation}
\begin{equation}
    \Sigma_a = \text{diag}(d_1,\dots,d_i), \text{ }d_i = \arg\min_{d_i} \left\| W_s^i - d_i \cdot U_a^i \right\|_2^2
    .
\label{approxi3}
\end{equation}
$U_s$ and $V_s$ are unitary matrices in the SVD form of $W_s$ as shown in (\ref{SVD}) and $W_s^i$ represents the $i$-th row of $W_s$. By solving a least square optimization problem for $W_s^i$, the $i$-th element of the diagonal matrix $\Sigma_a$, $d_i$, can be determined. With one unitary matrix eliminated, the area cost for each square matrix can be reduced by nearly 50\%.



To reduce the area cost of the ONN while maintaining the accuracy, weight matrices in selected ONN layers are partitioned as described in Fig.~\ref{matrix partition} and approximated using (\ref{approxi1}), which is referred to as matrix approximation. However, the matrix approximation can introduce errors. Therefore, we apply a hardware-aware training algorithm to maintain the accuracy.
When training the network, the loss is defined as the averaged weighted Mean Square Error (MSE) on the raw outputs for the first $E_1$ epochs. After $E_1$ epochs, the training is finetuned by directly applying the averaged MSE of the reconstructed gradients.
The loss function is determined as a two-stage function as follows:
\begin{equation}
    \mathcal{L} =
    \begin{cases}
    \displaystyle
    \frac{1}{|D|} \sum_{d=1}^{|D|} \sum_{i=1}^{M} W_T^{(i)} \left( O_{d,i} - O_{d,i}^* \right)^2, & \text{if } E < E_1 \\
    \displaystyle
    \frac{1}{|D|} \sum_{d=1}^{|D|} \left(\overline{G}_d - \overline{G}_d^*\right)^2, & \text{otherwise}
    \end{cases}
\end{equation}
where $d$ denotes the $d$-th data sample in the dataset $D$ and $E$ denotes the current epoch. $O_{d,i}$ and $O_{d,i}^*$ represent the received outputs and the expected outputs. $\overline{G}_d$ and $\overline{G}_d^*$ are the reconstructed gradient from the received signals and the expected quantized averaged gradient, respectively. $W_T$ represents the importance of the output bits chosen for training. 


During training, to guide the weight matrices in the selected layers toward the structure defined in (\ref{approxi1}), the matrix approximation algorithm is applied periodically.
After the training, if the approximation is not applied at the last epoch, it is enforced on the selected layers to ensure that the trained network matches the ONN structure.

\subsection{Scalability of the OptINC Architecture}
\label{Scalable Architecture}

\begin{figure}[!t]
\captionsetup{skip=8pt}
\centerline{\includegraphics[scale=0.9]{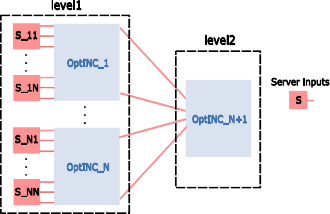}}
\small
\captionsetup{font=small}
\caption{\textbf{The cascading topology with OptINCs in two levels, supporting up to $N^2$ servers.}}
\vspace{-20pt}
\label{scalable}
\end{figure}

As discussed in Section~\ref{Optical INC Architecture}, the input size of the ONN in the OptINC architecture depends on the number of supported servers $N$. As $N$ increases, both the ONN and the required datasets scale accordingly, which makes training challenging. In this section, we demonstrate that a fixed OptINC architecture can be adapted to efficiently support a larger number of servers.

Fig.~\ref{scalable} illustrates a cascading OptINC topology that supports 
up to $N^2$ servers, with  $N$ OptINCs in level 1 and one OptINC in level 2. 
For fewer servers, unused inputs to the 
OptINCs can be connected to zero input, and completely unused OptINCs can be removed.
%
However, the cascading topology based on basic OptINCs can introduce errors due to two-level quantization. The expected averaged gradient $\overline{G}^*$ and the obtained result $\overline{G}_{basic}$ under two-level quantization are given by:
\begin{equation}
    \overline{G}^* = \textbf{Q }(\frac{1}{N^2}\sum_{i=1}^{N}\sum_{j=1}^{N}G_{i,n})
\end{equation}
\begin{equation}
    \overline{G}_{basic} = \textbf{Q }(\frac{1}{N}\sum_{i=1}^{N}\textbf{Q }(\frac{1}{N}\sum_{j=1}^{N}G_{i,n}))
\label{two-level}
\end{equation}
where $G_{i,n}$ denotes the gradient of server $n$  connected to the $i$-th OptINC. In (\ref{two-level}), the decimal parts of the averaged gradients are discarded during quantization in level 1, which can accumulate across levels and result in accuracy loss.

To avoid this error, when creating the training datasets, we keep the discarded decimal parts $d$ in level 1 as:
\begin{equation}\label{eqd}
    \overline{G}_{new} = \textbf{Q }(\frac{1}{N}\sum_{i=1}^{N}(\textbf{Q }(\frac{1}{N}\sum_{j=1}^{N}G_{i,n})+d)).
\end{equation}
By using these adapted datasets to train the ONNs in the OptINCs in the cascading topology, the actual averaged gradient 
$\overline{G}_{new}$ can then become equivalent to $\overline{G}^*$. 

\begin{table*}[!t]
\small
\captionsetup{font=small}
\caption{\textbf{Experimental results under different scenarios with various bit widths and server counts.}}
\begin{center}
\begin{tabular}{c c c c c c}
\toprule
\textbf{Bit Width} & \textbf{\#Servers} & \textbf{ONN Structure} & \textbf{Layers With Matrix Approximation} & \textbf{Area Ratio} & \textbf{ONN Accuracy} \\
\midrule
\multirow{2}{*}{\textbf{8}} & \multirow{2}{*}{4} & \multirow{2}{*}{\makecell{4-64-128-256-128-64-4}} & None  & 100\% & 100\% \\
 &   &  & All layers & 39.3\% & 100\% \\
 \midrule
\multirow{2}{*}{\textbf{8}} & \multirow{2}{*}{8} & \multirow{2}{*}{\makecell{4-64-128-256-512-\\256-128-64-4}} & None & 100\% & 100\% \\
 &   &  & Layers 2--7 & 40.9\% & 100\% \\
 \midrule
\multirow{2}{*}{\textbf{8}} & \multirow{2}{*}{16} & \multirow{2}{*}{\makecell{4-64-128-256-512-1024-\\512-256-128-64-4}} & None & 100\% & 100\% \\
 &   &  & Layers 2--9 & 40.4\% & 100\% \\
 \midrule
\multirow{2}{*}{\textbf{16}} & \multirow{2}{*}{4} & \multirow{2}{*}{\makecell{4-64-128-256-512-\\256-128-64-8}} & None & 100\% & 100\% \\
 &   &  & Layers 4--6 & 49.3\% & 100\% \\
\bottomrule
\end{tabular}
\label{table1}
\end{center}
\vspace{-20pt}
\end{table*} 

The decimal part $d$ in (\ref{eqd}) is merged into the last PAM4 output signal of the corresponding OptINC in level 1 and propagated to the OptINC in level 2, increasing the signal resolution at both levels. Therefore, a larger ONN is adopted to maintain the computation accuracy. Notably, OptINCs in both levels share the same expanded ONN structure, with each level trained on its modified dataset.



\section{Experimental Results}
\label{experimental results}
\begin{figure}[!t]
\captionsetup{skip=8pt}
\centerline{\includegraphics{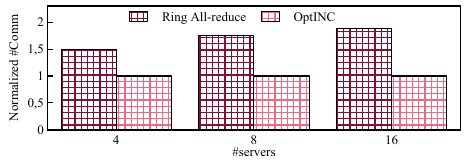}}
\small
\captionsetup{font=small}
\caption{\textbf{The communication data normalized by the amount of data to be computed, for the ring all-reduce algorithm and OptINC, when 4, 8, 16 servers participate in the distributed learning.}}
\vspace{-20pt}
\label{comm}
\end{figure}

To evaluate the proposed architecture, four distributed learning scenarios with various bit width and server number combinations were considered. For each scenario, a dedicated neural network was trained and mapped to the ONN in the OptINC architecture. We assumed that the weight matrices are mapped onto interleaving MZI arrays as in~\cite{onn1}, and that the nonlinear functions are implemented as in~\cite{nonlinear2}.  For simplicity, a Multilayer Perceptron (MLP) with ReLU activations was employed in all scenarios. The MLPs were trained with PyTorch using NVIDIA A100 Tensor Core GPUs.

Table~\ref{table1} shows the experimental results of the four scenarios. 
The first and the second columns list the bit widths and the number of supported servers, respectively. The third column specifies the ONN structures as a list of the numbers of neurons in the layers. 
The fourth column lists the selected layers to apply matrix approximation. The area ratio compared to the OptINC architecture without matrix approximation in (\ref{approxi1}) is shown in the fifth column, where the area cost is defined as the number of MZIs required to map the matrices in the OptINC architecture. The last column lists the trained accuracy of the corresponding ONN. When matrix approximation is applied, the OptINC area cost can be reduced to 39.2\%–49.3\% of that of the OptINC without approximation. With the hardware-aware training described in Section~\ref{training}, the accuracy can still be preserved at 100\%.

The output size of the ONNs was determined as the number of output PAM4 signals to represent the averaged gradients 
and the input size of the ONNs was set to four to balance the data complexity and training efficiency. 
The network depths and layer dimensions were selected through a greedy search, which in future work can be optimized by using Neural Architecture Search (NAS) algorithms.

Since OptINC offloads all computations to the network, incurring no extra rounds to exchange the data, the excessive communication data in the ring all-reduce algorithm can be eliminated. Fig.~\ref{comm} illustrates the communication data for 4, 8, 16 servers when the ring all-reduce algorithm and OptINC are used. The communication data is normalized by the gradients to be averaged. The ring all-reduce algorithm requires excessive communication data transfer, as described in Section~\ref{inrtoduction}, where the $\frac{N-2}{N}$ communication overhead ranges from $50\%$ to $87.5\%$. OptINC offloads the gradient computation to the network, where gradients are averaged once the signals traverse the network, eliminating the communication overhead. The corresponding improved latency performance, which also depends on the trained models and hardware specifications, is discussed later. 

\begin{table}[!t]
\small
\centering
\setlength{\tabcolsep}{3pt}
\captionsetup{skip=5pt, font=small}
\caption{\textbf{Selected layers for matrix approximation and the training accuracy in the fourth scenario.}}
\begin{tabularx}{\columnwidth}{c c X c} 
\toprule
\textbf{Layers} & 
\makecell{\textbf{ONN} \\ \textbf{Acc. (\%)}} & 
\centering \textbf{Error Values (Rel. Ratios \%)} & 
\makecell{\textbf{Norm.} \\ \textbf{Area}} \\
\midrule
4, 5, 6 & 100 & \centering None & 49.3\% \\
4, 5, 6, 7 & 99.99986 & \centering $\pm 1$ (90), $-64$ (10) & 47.9\% \\
4, 5, 6, 7, 8 & 99.99999 & \centering 1024 (100) & 47.4\% \\
3, 4, 5, 6 & 99.98891 & \centering $\pm 1$ (99), $\pm 1024$ (0.9), \\ $-4$ (0.1) & 43.7\% \\
3, 4, 5, 6, 7 & 99.99936 & \centering $\pm 4$ (79.5), $-16$ (17), \\ 12 (3.5) & 42.2\% \\
\bottomrule
\end{tabularx}
\label{table2}
\vspace{-20pt}
\end{table}

We explored the performance of the ONN using the fourth scenario in Table~\ref{table1} with different configurations of matrix approximation. The results are listed in Table~\ref{table2}, where the first column specifies the layers selected for matrix approximation. The accuracy of the trained ONNs is shown in the second column, while the introduced errors with their relative ratios are listed in the third column. For example, $\pm$1(90\%) indicates that, if the ONN fails to achieve 100\% accuracy and then introduces errors to the averaged gradient, the errors equal $\pm1$ in 90\% of the cases.
The corresponding reduced area costs are listed in the last column, normalized to the original OptINC architecture without approximation. As more large layers adopt matrix approximation, the area cost can be further reduced from 49.3\% to 42.2\%, at the expense of introducing errors with small probabilities.

We evaluated the impact of the proposed OptINC architecture on real distributed training tasks by simulating the training of two models with the OptINC architecture in Table~\ref{table2}. During training, the introduced errors in the third column of Table~\ref{table2} with their corresponding probabilities were injected into the averaged gradients. Specifically, ResNet50~\cite{resnet} was trained from scratch on the CIFAR-100 dataset for 300 epochs and a LLaMA-based network~\cite{llama2} with 8 layers, each with a hidden dimension of 384 and 8 attention heads, was trained on the Wikipedia-1B dataset for 50,000 steps. Floating-point gradients were quantized to fixed-point values using a global block quantization scheme similar to~\cite{inc2}, incurring a negligible synchronization cost of less than 0.4\% for both models.

\begin{figure}[!t]
\centering
\begin{tikzpicture}
\node[anchor=south west,inner sep=0] (image) at (0,0) {\includegraphics{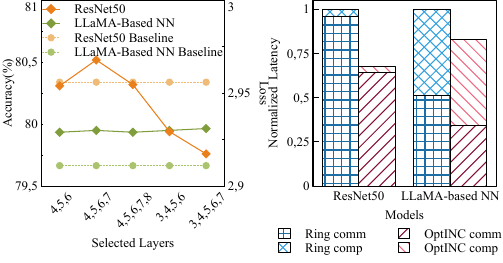}};
\begin{scope}[x={(image.south east)},y={(image.north west)}]
  \node[anchor=north west, fill=white] at (0.23,0.005) {(a)};
  \node[anchor=north east, fill=white] at (0.82,0.005) {(b)};
\end{scope}
\end{tikzpicture}
\small
\captionsetup{skip=8pt}
\captionsetup{font=small}
\caption{\textbf{(a) The trained accuracy and loss, and (b) the overall latency breakdown for one epoch or step of ResNet50 on CIFAR-100 and the LLaMA-based network on Wikipedia-1B.}}
\vspace{-20pt}
\label{results}
\end{figure}

 The training results are shown in Fig.~\ref{results}(a). For comparison, baselines were obtained by simulating accurate gradient averaging in servers for the ring all-reduce algorithm. Without error injection, both ResNet50 and the LLaMA-based network achieved comparable results to the baseline, with only a slight accuracy drop of 0.03\% on CIFAR-100 and a loss increase of 0.018 on Wikipedia-1B due to the block quantization. With error injection, the accuracy of ResNet50 slightly decreased by 0.55\%, while the loss of the LLaMA-based network increased by 0.02. For both models, the accuracy and loss still remained within acceptable ranges. 

Fig.~\ref{results}(b) illustrates the modeled latency breakdown when training the models for one epoch or step. The latency is normalized by the overall latency with the ring all-reduce algorithm. The setting included Nvidia H100 GPUs with a compute capability of 60 TFLOPs~\cite{h100}, a utilization efficiency of 0.6, and eight full-duplex transceivers, each of which has a bandwidth of 800 Gb/s~\cite{pam4}. For ResNet50, a convolutional neural network with less intensive computation workload than the transformer architecture, the communication latency dominated and OptINC reduced the overall latency by over 25\%. For the LLaMA-based network, where the latency for computation and communication were comparable, OptINC reduced the overall latency by around 17\%. Since this scenario only supported four servers, the latency improvement would show an increasing trend when supporting more servers, as shown in Fig.~\ref{comm}.

Finally, a scalable architecture was validated under the first scenario, where one OptINC supports four servers. By cascading five such OptINCs in two levels as in Fig.~\ref{scalable}, up to sixteen servers can be supported. 
To accommodate the increased resolution of the averaged gradients, 
the ONN structure listed in Table~\ref{table1} was modified by inserting two $64\times 64$ weight matrices with matrix approximation after the first layer and before the last layer, respectively, while other layers remained unchanged. 
For both levels, the ONNs in the OptINC architecture can be trained to reach 100\% accuracy on the modified dataset. Compared with the ONN structures in Table~\ref{table1}, this modification incurred about 10.5\% hardware overhead. 


\section{Conclusion}
\label{conclusion}
In conclusion, a scalable OptINC architecture was proposed to eliminate the communication overhead of the ring all-reduce algorithm by offloading the computation onto the network. 
Specifically, an ONN was employed to map the encoded gradients from multiple servers to the quantized averaged gradients. 
To reduce the dataset complexity, a preprocessing unit and a splitting unit were introduced. 
To decrease the hardware cost, selected weight matrices were partitioned and approximated with unitary matrices, leading to an area reduction of nearly 50\%.
The accuracy of the proposed architecture can still be maintained by a hardware-aware training algorithm. 
Future work will address physical-layer non-idealities and explore different network topologies and protocols.

\bibliographystyle{ieeetr}
\footnotesize
\bibliography{References}

\end{document}